# Evaluating LLM-Based Translation of a Low-Resource Technical Language: The Medical and Philosophical Greek of Galen

James L. Zainaldin[a,b,1], Cameron Pattison[c,d], Manuela Marai[b], Jacob Wu[e], and Mark J. Schiefsky[b,f]

[a]Department of Classical and Mediterranean Studies, Vanderbilt University, Nashville, TN 37235, USA; [b]Center for Hellenic Studies, Harvard University, Washington, DC 20008, USA; [c]Department of Philosophy, Vanderbilt University, Nashville, TN 37235, USA; [d]Machine Intelligence and Normative Theory (MINT) Lab, Australian National University, Canberra, ACT 2601, Australia; [e]Independent Researcher, San Diego, CA 92127, USA; [f]Department of the Classics, Harvard University, Cambridge, MA 02138, USA

[1]To whom correspondence may be addressed: james.zainaldin@vanderbilt.edu

## Abstract

*Purpose:* This study evaluates the quality of commercial large language model (LLM) machine translation (MT) for Ancient Greek technical prose and benchmarks standard automated MT evaluation metrics against expert human judgment.

*Design:* We evaluated 60 translations by three LLMs (ChatGPT, Claude, Gemini) of 20 paragraph-length passages from 2 works by the Greek physician Galen (c. 129–216 CE): an expository text with two published English translations and a pharmacological text never before translated. Quality was assessed using seven automated metrics and systematic reference-free human evaluation via a modified Multidimensional Quality Metrics (MQM) framework applied by domain specialists.

*Findings:* On the translated expository text, LLMs achieved high quality (mean MQM score 95.2/100). On the untranslated pharmacological text, quality was lower (79.9/100) but bimodally distributed: two passages with extreme terminological density produced catastrophic failures, while remaining passages scored within 4 points of the expository text. Terminology rarity, operationalized via corpus frequency, emerged as the dominant predictor of failure ($r = -.97$). Automated metrics showed moderate correlation with human judgment only on texts with wide quality variance; no metric discriminated among high-quality translations.

*Originality:* This is the first systematic, reference-free expert human evaluation of LLM translation for any ancient language and the first study identifying textual properties predictive of translation failure.

*Contribution to digital humanities:* Our findings establish a replicable methodology for evaluating LLM performance on specialized historical texts and demonstrate that corpus frequency analysis can flag content likely to contain critical translation errors—a scalable heuristic applicable across low-resource languages and specialized domains.

**Author Contributions** J.L.Z. designed research and led the project; J.L.Z., C. P., and M.S. conceptualized the study; C.P. developed the computational pipeline and contributed analytic tools; M.M. prepared expert reference translations and performed MQM evaluation, with assistance from J. L. Z.; J.W. curated data; J.L.Z. and C.P. analyzed data; and J.L.Z. wrote the paper with contributions from M.M. and feedback from C.P., J.W., and M.S.

**Competing Interest Statement:** The authors declare no competing interest.

**Data Availability:** All data, code, and materials used in the analysis are available at https://github.com/campattison/galen_project.

**AI Disclosure:** The authors used Anthropic's Claude 4.5/4.6 Sonnet and Opus and Google's Gemini 3.0/3.1 Pro to assist with manuscript editing and code development; the authors assume full responsibility for the accuracy of all content.

**Funding Statement:** This research was supported in part by the Center for Hellenic Studies, Harvard University (M.S., Director). J.L.Z. funded LLM API costs, web hosting, and data processing with personal research funds from Vanderbilt University.

## 1 Introduction

Large language models are increasingly deployed for tasks requiring specialized domain knowledge, yet the boundaries of their competence remain poorly characterized. When do these models produce reliable expert-level output, and when do they fail? We address this question through a demanding test case: translation of Ancient Greek technical prose, a task combining the challenges of low-resource languages with dense specialist terminology whose correct rendering requires domain expertise. Using expert human evaluation of 60 machine translations from the Greek physician Galen of Pergamum (c. 129–216 CE), we show that LLM performance achieves impressive results for expository prose but collapses as the density of rare terminology in the source text increases.

Low-resource ancient languages (LRAL) such as Ancient Greek (AG) have long challenged machine translation (MT). Despite incremental improvements from statistical to neural MT, systems continue to struggle with languages lacking large parallel corpora (Karakanta et al. 2018; Magueresse et al. 2020; Haddow et al. 2022; Sommerschield et al. 2023; Tafa et al. 2025), and inconsistent performance has impeded operationalization in fields such as Classics (Ross 2023; Tzanoulinou et al. 2025). Hyper-large language models (LLMs) such as Anthropic's Claude, Google's Gemini, and OpenAI's ChatGPT may change this. These models are widely accessible (Robinson et al. 2023; Volk et al. 2024; Tzanoulinou et al. 2025) and demonstrate impressive emergent cross-lingual capabilities (Wei et al. 2022; Muennighoff et al. 2023; Zhu et al. 2024; Akavarapu et al.

2025; Qin et al. 2025). While LLMs have struggled with many under-resourced languages (Hendy et al. 2023; Court and Elsner 2024; Shu et al. 2024; Ataman et al. 2025), early results show striking improvement for Classical languages including Latin and AG over last-generation neural MT (Gutherz et al. 2023; Volk et al. 2024; Wannaz and Miyagawa 2024; Akavarapu et al. 2025). These developments prompt the central question of this study: how good are commercial LLMs at translating AG, where "how good" is understood as "how successful by the lights of disciplinary norms and as judged by domain-specific expertise"?

The stakes are applicable to but extend far beyond Classical scholarship. As LLMs are adopted across various domains—from legal analysis to clinical medicine to materials science—practitioners face the same fundamental dilemma. LRAL offers an unusually rigorous setting for studying this problem. Characterizing LLM performance for AG technical prose could help guide scholars between the Scylla of eschewing these tools entirely and the Charybdis of employing them without adequate understanding of their strengths and weaknesses—a tension familiar to any field where AI is outpacing trust calibration. Understanding how well commercially available LLMs translate AG "out of the box" could both legitimate a range of use cases immediately available to the scholarly community and establish a methodological template for assessing LLM reliability in other specialized domains.

### 1.1 Challenges in Evaluation of MT for AG

Evaluating MT quality for AG is deeply problematic for reasons both of its intrinsic linguistic complexity and the limitations of current methods. The language is morphologically rich, syntactically flexible, and diachronically varied, with texts often transmitted in corrupt form requiring expert judgment to establish a reliable reading (Smyth 1920; West 1973; Horrocks 2010; Reynolds and Wilson 2014; van Emde Boas et al. 2019). These properties augment an already difficult evaluation landscape: expert human translators regularly disagree on lexical choices and discourse-level interpretation (Mendelsohn 2011), translations vary by purpose and audience (Venuti 2008; Bassnett 2013), and many texts lack modern translations entirely. This interpretive plurality makes reference-based automated evaluation—the standard approach in MT research (Papineni et al. 2002 for BLEU; Celikyilmaz et al. 2021; Sai et al. 2020)—especially unreliable for AG, since any given reference may represent only one of several defensible interpretations (Hardwick 2000; Hardwick and Stray 2008; Lianeri and Zajko 2008). Standard automated metrics face further compounding difficulties: lexical metrics such as BLEU penalize valid paraphrases in a morphologically rich language (Callison-Burch et al. 2006; Reiter 2018; Freitag et al. 2022; Papantoniou and Tzitzikas 2024; Tzanoulinou et al. 2025), embedding-based metrics such as BERTScore (Zhang et al. 2020) lack syntactic sensitivity, and learned neural metrics such as COMET (Rei et al. 2020) trained on modern data may reward fluent hallucinations over faithful renderings (Amrhein and Sennrich 2022; Guerreiro et al. 2023b). (Emerging LLM-as-judge approaches, for which see Fernandes et al. 2023; Kocmi and Federmann 2023, pose intriguing possibilities but require further evaluation, which we reserve for a separate study.) These limitations motivate our use of systematic, reference-free human evaluation.

A growing body of work has applied AI and LLMs to Classical languages (Berti 2019; Graziosi et al. 2023; Riemenschneider and Frank 2023, Sommerschield et al. 2023; Stopponi et al. 2024; Tzanoulinou et al. 2025), including evaluations of LLM-based translation for AG and other LRAL including Akkadian, Coptic, and Latin (Gutherz et al. 2023; Volk et al. 2024; Wannaz and Miyagawa 2024; Rosu 2025; Zhuang 2025). These studies have shown that LLMs can outperform older MT platforms on standard automated metrics and have compared models against one another. Some studies have further utilized domain-specific human expertise to "spot-check" translations (Akkadian: Gutherz et al. 2023) or to assess accuracy of themes and content (post-antique Latin: Volk et al. 2024), but none has attempted systematic, reference-free human evaluation; none has involved human evaluation of AG MT; and no study has identified the textual properties that predict translation failure or established the empirical boundary between reliable and unreliable LLM output for any ancient language. Additionally, nearly all prior work has evaluated MT only for texts with existing modern translations, compelled by the need for a reference translation for automated evaluation. Gutherz et al. 2023 exceptionally considered three untranslated texts and Volk et al. 2024 evaluated summaries rather than direct translations, but no study has systematically compared LLM performance on translated versus untranslated texts. This is a major desideratum given that LLM-based MT might be contaminated by human translations in their training data (Carlini et al. 2021; Chang et al. 2023; Sainz et al. 2023; Yang et al. 2023; Deng et al. 2024; Ji et al. 2024; Xu et al. 2024): can LLMs really translate *de novo* or are they really only regurgitating human translations present in their training data?

### 1.2 A Case Study in Evaluating LLM-Based MT: Galenic Medicine

This paper explores the question of the efficacy of LLMs in translating AG by bringing human domain expertise into the loop. We entertain two principal aims:

(1) to directly measure the quality of LLM-based MT from AG to English on both previously translated and previously untranslated texts, using not only automated MT evaluation metrics but also expert human judgment via a modified Multidimensional Quality Metrics (MQM) framework;
(2) to benchmark the standard automated MT translation evaluation metrics against expert human judgment of translation quality to determine whether they offer any meaningful insight into the quality of LLM-based MT.

We address this question through the presentation of a subfield case study drawn from the discipline of Classics: translation of AG medical, philosophical, and scientific writings, specifically, those of the Greek physician Galen of Pergamum (c. 129–216 CE).

This case is compelling for multiple reasons. First, it offers well-defined quality criteria, since successful translation of AG technical writings can be understood mainly as the accurate translation of the text's *concepts* and *claims*, abstracting from *rhetorical* or *aesthetic* elements. This simplifies, but our translation goal follows from sociolinguistic research on "technical languages," whose features are oriented towards the communication of extra-textual expertise (Hoffmann et al. 1998; Langslow 2000; Fögen 2009; Zainaldin 2020). Second, Galen's voluminous corpus contains both works that have been translated into English and those never translated (Fichtner 2023). This intra-corpus variance allows us to assess the performance of LLM-based MT on both translated and untranslated AG. Third, the internal variety of Galen's writing creates a natural gradient of terminological difficulty within a single author's corpus, potentially improving generalizability to the broader body of AG medical, philosophical, and scientific writings.

Our paper mobilizes domain-specific expertise in Classics and history of medicine, philosophy, and science to evaluate LLM-based MT quality and benchmark standard automated evaluation metrics against expert human judgment. We established a dataset of 60 LLM translations (Claude, Gemini, ChatGPT) of paragraph-length AG excerpts from two Galenic works—one with two published English translations, one never translated—and evaluated them using both a suite of automated metrics and systematic, reference-free human evaluation via a modified MQM framework. We report results in aggregate and by model, comparing translated versus untranslated texts, stratifying by passage-level difficulty, and benchmarking automated metrics against human judgment. We conclude with discussion of practical implications and avenues for further work.

## 2. Data

### 2.1 Galen of Pergamum

Galen of Pergamum (c. 129–216 CE), along with Hippocrates of Cos (circa 5th c. BCE), is the most important medical author of Greek and Roman antiquity, responsible for approximately 10% of all AG literature down to 350 CE (Nutton 2020, Nutton 2023). His vast corpus spans medical treatises, philosophical works, commentaries, philological texts, and autobiographical writings (Boudon-Millot 2007; Hankinson 2008; Mattern 2013; Nutton 2020). This diversity and his central position in ancient traditions of science, medicine, and philosophy make him an ideal candidate for testing LLM-based MT from AG.

### 2.2 The AG Dataset

We established the AG dataset for this study by excerpting two of Galen's works: (1) *On Mixtures* (Περὶ κράσεων = *De temperamentis*), hereafter "*Mix.*," a work of approximately 28,000 words in three parts systematically elaborating Hippocratic humoral theory into a framework of biological qualities; (2) *On the Composition of Drugs according to Kinds* (Περὶ συνθέσεως φαρμάκων = *De compositione per genera*), hereafter "*Comp.*", a work of approximately 102,000 words in seven parts, a pharmacological recipe book with thousands of formulations for topical applications, mostly wound healing plasters, as well as multipurpose and emollient plasters. These texts represent a meaningful spectrum of Galenic writing, from the expository *Mix.*, which relates philosophical reasoning to medicine and natural science, to the highly technical *Comp.*, with its dense lists of procedures and ingredients (many not securely identified today). We used the best available scholarly editions: G. Helmreich (Helmreich 1904) for *Mix.* and C. G. Kühn for

*Comp.* (Kühn 1827). Kühn's text is of inferior quality and contains errors, but is generally readable and has not been replaced (Nutton 2002). A further motivation in selecting these works is that *Mix.* has two recent English translations (Singer and van der Eijk 2018; Johnston 2020), while *Comp.* has never been translated into English, enabling a comparative test.

To establish our dataset of AG passages, we extracted representative passages from across the texts. For each work, we excerpted a paragraph-length passage of 215 words (*Mix.*) and 214 words (*Comp.*) from the beginning. We then segmented each work into 10 approximately equal divisions by word count and excerpted the initial passage from each division as close to the initial segment length. This procedure yielded 20 total passages of similar length, which are detailed in the Supplementary Materials (Table SM1). Previous studies evaluating MT of LRAL have focused on sentence-length input, but we selected paragraph-length passages on the supposition that longer inputs better model real-world use. Testing input-length effects was not practicable (Liu et al. 2023; Shi et al. 2023; Pang et al. 2025). Philological inspection by expert team members confirmed that the extracted passages were reasonably representative cross-sections of each work.

### 2.3 Reference Translations

We synthesized an English reference-translation dataset corresponding to the AG passages. For *Mix.*, we used the translations of P. N. Singer and P. J. van der Eijk (2018), published by Cambridge University Press, and I. Johnston (2020), published in the Loeb Classical Library by Harvard University Press. Despite significant differences between translations, we regarded peer-reviewed publication by major academic university press as sufficient imprimatur for gold-standard references. Since no reference translation exists for *Comp.*, a team member (M. Marai) with scholarly expertise in this text prepared English translations, revised in consultation with other expert team members (M. Schiefsky, J. Zainaldin).

### 2.4 Machine Translations

To produce English MT passages, we used three commercial LLMs: Anthropic's Claude (claude-sonnet-4-5-20250929), OpenAI's ChatGPT (gpt-5-2025-08-07), and Google Gemini (gemini-2.5-pro). Our automated translation pipeline is published online.[i] We used a standardized prompt (Appendix A1) that exceeds the bare "translate" commands of some past studies but does not optimize for the technical content. We produced 60 total translations (20 passages × 3 models), which were blinded and randomly ordered before evaluation.

## 3. Methodology

To evaluate LLM-based MT performance, we implemented both automated (reference-based) and expert human (reference-free) frameworks. The automated tests are cheap, fast, and widely standardized, but of uncertain efficacy. We therefore undertook domain-expert examination of all 60 MT translations via modified MQM evaluation, which allowed us to capture granular, explanatory knowledge of the performance of the LLMs with rich data on both error type and severity.

### 3.1 Automated MT evaluation

We obtained MT translation evaluation scores from our dataset using standard lexical, embedding-based (semantic), and learned neural metrics. We included a variety of automated metrics not only to reproduce standard methods but also to furnish a range of metrics for benchmarking against human evaluation. Because *Mix*. possesses two published reference translations while *Comp*. required a newly created reference (section 2.3), multi-reference protocols applied only to *Mix*. Our code pipeline for automated translation evaluation is also published online.

#### 3.1.1 Lexical metrics

We employed four standard lexical metrics: **BLEU-4** (Papineni et al. 2002), which calculates the geometric mean of *n*-gram precisions (up to 4-grams) with a brevity penalty; **chrF++** (Popović 2017), which measures character- and word-level n-gram overlap to account for morphological variation; **METEOR** (Banerjee and Lavie 2005; Denkowski and Lavie 2014), which incorporates stemming and synonym matching to credit lexical equivalents beyond exact matches; and **ROUGE-L** (Lin 2004), which computes longest common subsequence overlap to capture sentence-level structure. For *Mix*., we utilized the standard multi-reference protocols native to each evaluation library.

#### 3.1.2 Embedding-based (semantic) metrics

We assessed semantic similarity using BERTScore (Zhang et al. 2020), which computes token-level cosine similarities from contextual embeddings, capturing semantic equivalence even when surface forms differ. For *Mix.*, we computed BERTScore against each reference independently and retained the maximum score for each passage.

#### 3.1.3 Learned neural metrics

We evaluated holistic translation quality using two learned metrics trained on human quality judgments: **COMET** (Rei et al. 2020), which employs a cross-lingual encoder to evaluate translations against both source and reference, and **BLEURT** (Sellam et al. 2020), a regression model fine-tuned to detect fluency and meaning errors. Because these models take a single reference as input, we performed separate inference passes for each *Mix.* reference and reported the maximum score.

### 3.2 Human MQM evaluation

We based our expert philological evaluation on Multidimensional Quality Metrics (MQM), a framework for translation quality assessment organized around error typologies and severity levels (Lommel et al. 2014; Mariana et al. 2015) that is the official evaluation strategy for the Workshop on Machine Translation (Freitag et al. 2021a; Freitag et al.

2021b). Previous studies evaluating MT for LRAL have relied on custom metrics (Volk et al. 2024) or qualitative spot-checks (Gutherz et al. 2023); none has applied the systematic analytical framework of MQM, in part because of its labor intensity. Nonetheless, MQM allows for fine-grained evaluation of translation quality utilizing domain-specific expertise. It does not presuppose the existence of authoritative reference translations, mitigating a central shortcoming of automated MT evaluation for AG: its dependence on unstable or incomplete reference data.

The MQM taxonomy is adaptable to ancient languages. Because our translation goals emphasize accurate representation of conceptual content rather than style, we defined a custom taxonomy around two dimensions: terminological fidelity and consistency and propositional content fidelity. We operationalized these translation goals respectively via the MQM taxonomies of (a) **Terminology**, with subcategories Terminological Accuracy and Terminological Consistency, and (b) **Accuracy**, with subcategories of Mistranslation, Overtranslation, Undertranslation, Addition, and Omission. We assumed the existence of hallucinations as a characteristic mode of failure for LLM-based MT (Lee et al. 2018; Guerreiro et al. 2023a; Guerreiro et al. 2023b; Ji et al. 2024) but did not operationalize a separate category because it is impracticably vague: translating AG often requires interpretative rendering even when accurate, leaving no pragmatic test distinguishing "hallucination" from other sorts of misconstrual, and fluent hallucinations always resolve as specific terminological or propositional errors.

Each MQM error was assigned a severity—Neutral, Minor, Major or Critical—with standard weighted multipliers (0-1-5-25) for penalty points. We computed a Translation Quality Score (TQS) for each passage by subtracting the weighted error severity count from 100, normalized by passage length (Freitag et al. 2021a; Lommel et al. 2024):

$$\text{TQS} = 100 - \left( \frac{(1 \times n_{\text{minor}}) + (5 \times n_{\text{major}}) + (25 \times n_{\text{critical}})}{\text{Word Count}} \times 100 \right)$$

To translate continuous TQS scores into discrete quality judgments and thus pragmatically evaluate the adequacy of the translations, we established three rating categories based on empirical inspection of score distributions and their correspondence to domain-specific comprehension. Scores ≥95 were graded as HIGH PASS, indicating adequacy for grasping the key claims and concepts of the Galenic text, even if not free of errors in all respects; 95 > TQS ≥ 87 were graded as LOW PASS, indicating that the MT could be used to grasp the gist of the Galenic text but might mislead or conceal on substantive points ; and scores < 87 were graded as FAIL, indicating sufficient quantity or severity of errors to render the translation unreliable without consulting the AG source.

We report these quality ratings under two schemes (section 4.4). Scheme 1 assigns ratings based on TQS alone. Scheme 2 applies an additional rule: reasoning that, for a subset of our translation goals, even a single critical error is individually sufficient to render an entire passage untrustworthy, we implemented a hard logic whereby any passage containing ≥1 critical error is automatically classified as FAIL regardless of TQS. The gated scheme reflects a conservative standard appropriate for users who cannot independently verify translations against the source text; the ungated scheme may better reflect utility for expert users capable of identifying and correcting isolated critical errors in an otherwise adequate translation.

Adapted MQM offers the best widely accepted framework for quantifying human evaluation of AG but does not resolve interpretative plurality, even given our narrow translation goals. Galenic terms can sustain multiple interpretations in context, including a general sense and a more specific technical shading, such as *energeia* (ἐνέργεια, *Mix.* 5) = "activity" or "function"; *pepsis* (πέψις, *Mix*. 5, 10, *Comp.* 8) = "concoction" or "digestion"; and so forth. Other terms lack consensus on their precise technical meaning, such as *chalkanthos* (χάλκανθος) and *chalchitis* (χαλκῖτις), copper compounds in *Comp.* 8, or *melilōton* (μελίλωτον), a clover in *Comp.* 10. Still other terms (pharmaceutical, pathological, and otherwise) pose challenges because of potential discrepancies between ancient and the present English medical meanings, such as *sandarache* (σανδαράκη, *Comp*, 8), *chrysocolla* (χρυσοκόλλα, *Comp*. 8), *melanthion* (μελάνθιον, *Comp*.10), and *nomai* (νομαί, *Comp*. 8). Finally, some words appear only once in the Greek language or a few times at most; the significance of these is often uncertain but can be conjectured from the root, such as with *proschoros* (πρόσχορος, *Comp.* 4) and *diaphorikos* (διαφορικός, *Comp.* 6). These may indeed be errors in the transmitted Greek text (although the correct reading could still challenge interpretation); such textual corruption was suspected elsewhere in Kühn's edition, such as the discussion of the spleen in *Comp.* 2 or the numbers in *Comp.* 9.

Both error type and severity are therefore subject to interpretation, given expected variability in expert judgment and uncertainties in how MT should be penalized for failing to translate corrupt or semantically uncertain terms. To mitigate this, expert team-members engaged in consensus-based decision-making for all 60 passages. Procedurally, at least two expert team members reviewed and tagged errors in every blinded MT passage, subsequently meeting to adjudicate discrepant error-types and severity weights to reach 100% consensus. Final MQM results thus represent a consensus of domain-specific expertise grounded in painstaking philological analysis of the entire Greek text and all 60 MT translations.

## 4. Results

### 4.1 Aggregate Translation Quality per Automated Metrics and MQM TQS

Table 1 presents automated MT evaluation scores for all 60 LLM translations, aggregated by text and model. Scores are reported for seven metrics: BLEU-4, chrF++, METEOR, and ROUGE-L (lexical); BERTScore (embedding-based); COMET and BLEURT (neural).

**Table 1. Aggregate Automated MT Evaluation Scores**

| *Text* | Model | BLEU-4 | chrF++ | METEOR | ROUGE-L | BERTScore | COMET | BLEURT |
|---|---|---|---|---|---|---|---|---|
| *Mix.* | ChatGPT | 31.4 (± 6.1) | 53.4 (± 3.7) | 46.4 (± 5.4) | 50.9 (± 5.3) | 91.0 (± 1.2) | 79.9 (± 1.9) | 49.8 (± 3.4) |

| | Claude | 34.2 (± 6.2) | 55.4 (± 3.4) | 48.5 (± 3.4) | 55.3 (± 5.6) | 91.6 (± 0.9) | 79.8 (± 2.1) | 50.4 (± 3.9) |
|---|---|---|---|---|---|---|---|---|
| | Gemini | 34.2 (± 5.0) | 57.0 (± 2.8) | 50.0 (± 3.9) | 56.0 (± 4.9) | 91.5 (± 1.0) | 80.7 (± 1.8) | 51.3 (± 4.3) |
| | ***Aggregate*** | 33.3 (± 5.7) | 55.3 (± 3.5) | 48.3 (± 4.4) | 54.1 (± 5.6) | 91.4 (± 1.0) | 80.1 (± 1.9) | 50.5 (± 3.8) |
| *Comp.* | ChatGPT | 15.7 (± 5.3) | 47.4 (± 5.2) | 40.1 (± 6.8) | 45.7 (± 6.2) | 89.1 (± 2.1) | 75.1 (± 4.0) | 42.6 (± 3.2) |
| | Claude | 16.7 (± 4.8) | 49.4 (± 2.4) | 42.9 (± 4.4) | 47.8 (± 2.7) | 89.7 (± 1.6) | 76.5 (± 2.3) | 46.2 (± 3.7) |
| | Gemini | 19.0 (± 4.0) | 51.2 (± 3.1) | 44.4 (± 4.7) | 47.8 (± 3.9) | 89.9 (± 1.3) | 77.3 (± 2.3) | 45.8 (± 6.2) |
| | ***Aggregate*** | 17.1 (± 4.8) | 49.3 (± 4.0) | 42.5 (± 5.5) | 47.1 (± 4.5) | 89.5 (± 1.7) | 76.3 (± 3.0) | 44.9 (± 4.7) |

*Note: All scores reported as mean (± SD) × 100.*

Among models, Gemini achieved the highest mean scores on most metrics for both texts, followed by Claude and ChatGPT, although inter-model differences were modest relative to standard deviations. Full per-passage results for all models for a selection of evaluation schemes appear in the Supplementary Material (Table SM2).

Table 2 presents MQM Translation Quality Scores (TQS) for all 60 LLM translations, aggregated by text and model. Full per-passage TQS are given in the Supplementary Material (Table SM3).

**Table 2. Aggregate MQM Translation Quality Scores**

| Text | Model | TQS Mean | TQS SD | Critical Errors |
|---|---|---|---|---|
| *Mix.* | ChatGPT | 92.3 | 6.6 | 4 |
| | Claude | 96.2 | 2.3 | 0 |
| | Gemini | 97.1 | 3.3 | 1 |
| | ***Aggregate*** | 95.2 | 4.8 | 5 |
| *Comp.* | ChatGPT | 74.9 | 32.7 | 24 |
| | Claude | 81.2 | 24.5 | 18 |
| | Gemini | 83.4 | 22.5 | 15 |
| | ***Aggregate*** | 79.9 | 26.2 | 57 |

*Note: TQS = Translation Quality Score (0–100). Critical Errors = count of errors rated “Critical” severity.*

As with the automated metrics, Gemini achieved the highest aggregate score on TQS for both texts, trailed closely by Claude (differences in aggregate from Gemini < 2) with ChatGPT in a distant third (differences in aggregate from Gemini of 4.8 and 8.5). In general, performance on *Mix.* was high, with mean aggregate TQS of 95.2 and 5 total

Critical errors across all 30 passages. Performance on *Comp.* was lower, with mean aggregate TQS of 79.9 and 57 total Critical errors across all 30 passages. While TQS for *Mix.* was thus approximately 15.3 points higher, far greater volatility for *Comp.* (SD = 26.2 in aggregate) versus *Mix.* (SD = 4.8 in aggregate) suggests a more complicated picture.

### 4.2 Passage-level Stratification of Translation Quality Metrics

Figure 1 visualizes model-performance for all 60 LLM translations by text and passage.

**Figure 1. MQM Translation Quality Scores by Work and Passage**

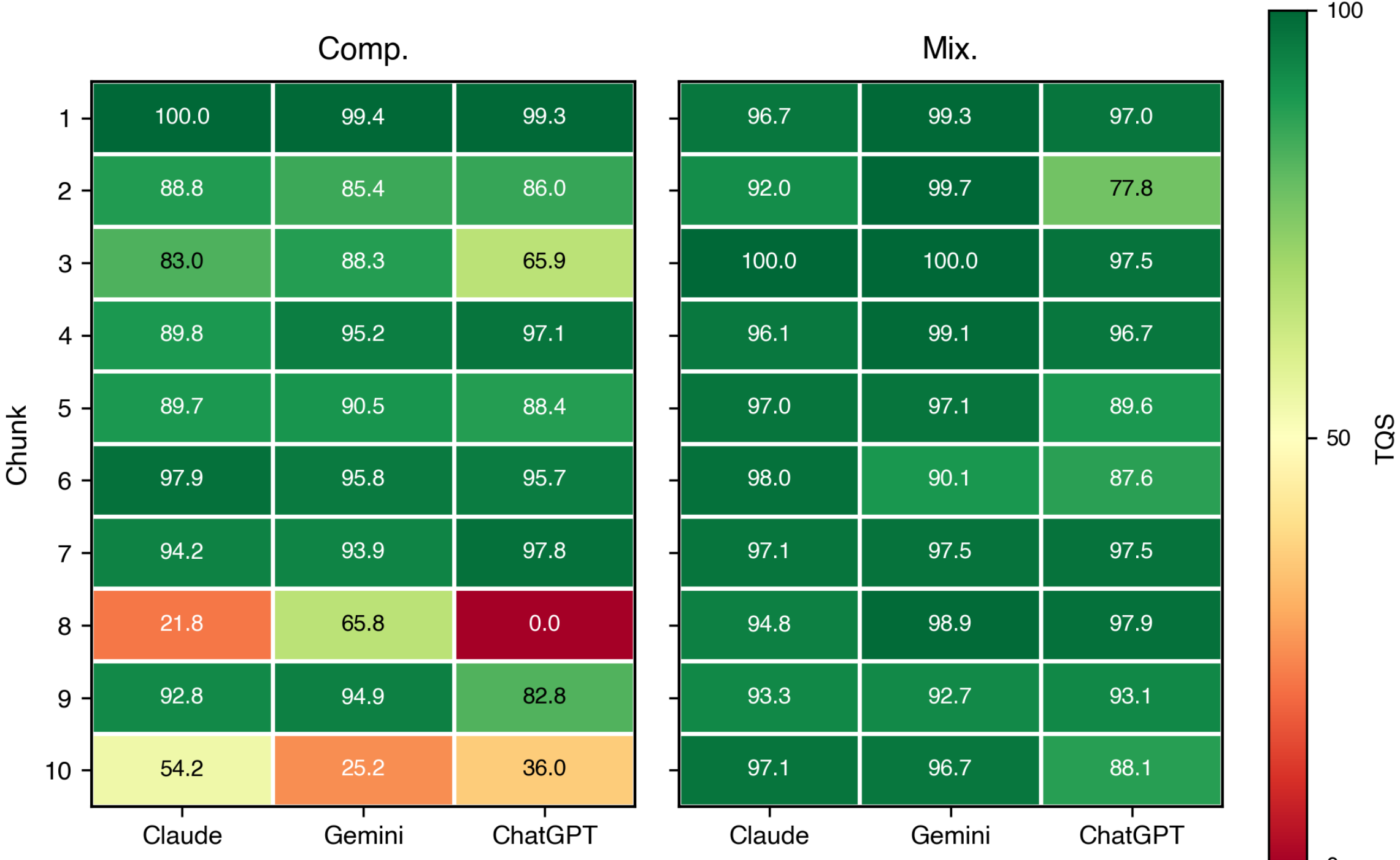

Alt text: Grouped bar chart showing MQM Translation Quality Scores (0–100) for three LLMs (Claude, Gemini, ChatGPT) across twenty passages from two Galenic works. The ten *Mix*. passages cluster tightly between approximately 78 and 100, with most scores above 90. The ten *Comp*. passages show greater variance: 1, 4, 6, 7, and 9 score comparably to *Mix*., while 8 and 10 produce catastrophic failures, with scores ranging from 0 (ChatGPT, passage 8) to approximately 66 (Gemini, passage 8). Passage 3 shows intermediate difficulty with higher inter-model variance.

As Figure 1 demonstrates, performance on *Comp.* was not uniformly poor but rather marked by variability. Two passages (8 and 10) produced catastrophic failures across all models: on *Comp*. 8, ChatGPT scored 0.0 (13 critical errors), Claude 21.8 (9 critical errors), and Gemini 65.8 (4 critical errors); on *Comp*. 10, scores ranged from 25.2 (Gemini) to 54.2 (Claude). Passage 3 also challenged all three models, but inter-model variance was significantly higher, with ChatGPT (65.9) dramatically trailing Gemini (88.3) and Claude (83.0). The remaining passages (1, 2, 4, 5, 6, 7, 9) yielded scores more comparable to *Mix*., ranging from 82.8 to 100.0 with no translations below 70, although passages 2 and 5 also posed difficulties (mean TQS 86.7 and 89.5, respectively).

Table 3 summarizes TQS by text and passage type for *Comp.* under several stratification schemes, taking into account passage-specific variability.

**Table 3. MQM Translation Quality Scores Stratified by Passage Exclusion**

| Stratification | Mean TQS | SD | Gap vs *Mix.* |
|---|---|---|---|
| *Mix.* (all passages) | 95.2 | 4.8 | — |
| *Comp.* (all passages) | 79.9 | 26.2 | -15.3 |
| *Comp.* (excl. 8, 10) | 91.4 | 7.5 | -3.8 |
| *Comp.* (excl. 3, 8, 10) | 93.1 | 5.0 | -2.1 |
| *Comp.* (excl. 8, 10, and ChatGPT on 3) | 92.5 | 5.3 | -2.7 |

*Note: Gap =* Mix. *mean TQS minus* Comp. *mean TQS under each stratification scheme.*

Excluding only the two universally catastrophic passages (8 and 10), *Comp*. scores rose to 91.4 (SD = 7.5), narrowing the gap with *Mix.* (15.3) to approximately 3.8 points. Further excluding passage 3 raised the *Comp*. mean to 93.1 (SD = 5.0), with a residual gap of 2.1 points. Finally, excluding ChatGPT's passage 3 result alone (while retaining Claude and Gemini's scores for that passage) yielded a *Comp*. mean of 92.5 (SD = 5.3) and a gap of 2.7 points. Under both specifications excluding the catastrophic passages 8 and 10, the gap between translated and untranslated texts was less than 4 points. Table 4 examines the *Mix.–Comp.* gap further on automated metrics under the same stratification schemes.

**Table 4. *Mix.–Comp*. Gap on Automated Metrics: Stratified Analysis**

| Metric | All passages | Excl. 8, 10 | Excl. 3, 8, 10 |
|---|---|---|---|
| BLEU-4 | −48.6% | −47.4% | −47.4% |
| chrF++ | −10.8% | −9.5% | −9.4% |
| METEOR | −12.0% | −9.7% | −8.7% |
| ROUGE-L | −12.8% | −12.9% | −13.8% |
| BERTScore | −2.0% | −1.3% | −1.4% |
| COMET | −4.8% | −3.9% | −3.4% |
| BLEURT | −11.1% | −11.1% | −11.5% |

*Note: Values represent relative performance drop from* Mix. *to* Comp. *(negative =* Comp. *scores lower). All aggregate gaps significant at* $p < .01$ *with large effect sizes (Cohen's* $d = .95$*–3.49, all passages).*

In aggregate, all metrics showed statistically significant gaps ($p < .01$) with large effect sizes ($d > 0.8$). Most metrics showed narrower relative declines, but BLEU-4 exhibited a massive collapse, dropping 48.6% from *Mix*. to *Comp*. ($p < .001$, $d = 3.06$). Unlike MQM scores, automated metrics continued to show substantial performance gaps even when catastrophic passages were excluded. BERTScore and COMET showed the smallest residual gaps (1.4% and 3.4% respectively, excluding 3, 8, and 10), yet these differences remained statistically detectable, while the residual BLEU gap remained very large (47.4%).

## 4.3 Correlation of Automated Translation Quality Evaluation Metrics and MQM

Table 5 demonstrates correlations between automated MT metrics and MQM TQS across all 60 translations.

**Table 5. Correlations between Automated Metrics and MQM TQS**

| Metric | Pearson *r* [95% CI] | *p*-value | Spearman ρ [95% CI] | *p*-value |
|---|---|---|---|---|
| BERTScore | +0.75*** [.62, .85] | < .001 | +0.43*** [.20, .62] | < .001 |
| COMET | +0.60*** [.41, .74] | < .001 | +0.51*** [.30, .68] | < .001 |
| METEOR | +0.53*** [.32, .69] | < .001 | +0.26* [.01, .48] | .044 |
| chrF++ | +0.53*** [.32, .69] | < .001 | +0.38** [.13, .57] | .003 |
| BLEU-4 | +0.45*** [.22, .63] | < .001 | +0.42*** [.18, .61] | < .001 |
| ROUGE-L | +0.34** [.10, 0.55] | .007 | +0.31* [.07, .53] | .014 |
| BLEURT | +0.32* [.07, 0.53] | .012 | +0.24 [-0.02, 0.47] | .065 |

*Note:* * p *< .05,* ** p *< .01,* *** p *< .001.* N *= 60 translations. 95% confidence intervals [in brackets] calculated via Fisher's* z*-transformation; all* p*-values are two-tailed.*

BERTScore showed the strongest linear association with TQS (Pearson $r = .75$, $p < .001$), followed by COMET ($r = .60$, $p < .001$), METEOR ($r = .53$, $p < .001$), and chrF++ ($r = .53$, $p < .001$). BLEU-4 ($r = .45$, $p < .001$) and ROUGE-L ($r = .34$, $p < .01$) showed weaker correlations. BLEURT, despite being a learned neural metric trained on human quality judgments, exhibited the weakest correlation ($r = .32$, $p = .012$), with a non-significant Spearman rank correlation ($\rho = .24$, $p = .065$). Spearman rank correlations were uniformly lower than Pearson correlations across all metrics, with COMET performing best ($\rho = .51$, $p < .001$) and METEOR showing the largest discrepancy between Pearson and Spearman values ($r = .53$ vs. $\rho = .26$). While this ordering is suggestive, confidence intervals for adjacent metrics overlap at this sample size, and the differences between individual metrics should be interpreted with caution. The moderate strength of the correlations indicate that automated metrics capture some dimensions of translation quality as assessed by domain experts but leave substantial variance unexplained—consistent with the divergent behavior of automated metrics and MQM scores observed in the stratified analysis.

## 4.4 Two Quality Evaluation Schemes for MQM

To translate continuous TQS scores into discrete quality judgments with practical significance, we applied the two rating schemes described in 3.2: Scheme 1 (TQS score thresholds only) and Scheme 2 (gated for ≥1 critical error = Fail). Table 6 presents quality ratings by text and model under both schemes.

**Table 6. Quality Ratings by Text and Model**

| Text | Model | Scheme 1 (TQS-only) | | | Scheme 2 (Gated) | | |
|---|---|---|---|---|---|---|---|
| | | High Pass | Low Pass | Fail | High Pass | Low Pass | Fail |
| *Mix.* | Claude | 70% | 30% | 0% | 70% | 30% | 0% |
| | Gemini | 80% | 20% | 0% | 80% | 10% | 10% |
| | ChatGPT | 50% | 40% | 10% | 50% | 20% | 30% |
| | **Aggregate** | 66.7% | 30.0% | 3.3% | 66.7% | 20.0% | 13.3% |
| *Comp.* | Claude | 20% | 50% | 30% | 20% | 20% | 60% |
| | Gemini | 30% | 40% | 30% | 30% | 20% | 50% |
| | ChatGPT | 40% | 10% | 50% | 40% | 0% | 60% |
| | **Aggregate** | 30.0% | 33.3% | 36.7% | 30.0% | 13.3% | 56.7% |

*Note: Scheme 1 assigns ratings based on TQS thresholds (High Pass ≥95, Low Pass 87–94, Fail <87). Scheme 2 applies an additional rule: any translation with ≥1 critical error is classified as Fail regardless of TQS. Percentages based on* n *= 10 translations per model per text (*n *= 30 per text aggregate).*

On *Mix.*, translation quality was high under both schemes. Under Scheme 1, 96.7% of translations achieved a passing rating (HIGH PASS or LOW PASS), with only 3.3% (1 of 30) classified as FAIL. Under Scheme 2, the pass rate remained strong at 86.7%, with the increase in failures (13.3%) attributable to 3 translations with TQS ≥ 87 but containing isolated critical errors. Claude achieved a 100% pass rate under both schemes; Gemini achieved 100% under Scheme 1 and 90% under Scheme 2; ChatGPT showed more variability, with pass rates of 90% (Scheme 1) and 70% (Scheme 2).

On *Comp.*, results diverged more sharply between schemes. Under Scheme 1, 63.3% of translations passed—a majority, though substantially lower than *Mix*. Under Scheme 2, the pass rate dropped to 43.3%. The difference between schemes reflects 6 *Comp.* translations with TQS in the 87–94 range (LOW PASS under Scheme 1) but containing critical errors (FAIL under Scheme 2). Among models, ChatGPT achieved the highest HIGH PASS rate on Comp. (40%) but also the highest FAIL rate under both schemes; Claude and Gemini were more balanced.

Table 7 presents quality ratings for *Comp.* under the same stratification schemes applied in section 4.2, allowing comparison of the TQS convergence findings with the quality rating analysis.

**Table 7. Quality Ratings Under Stratification**

| Stratification | Scheme 1 Pass Rate | Scheme 2 Pass Rate | Scheme 1 Gap | Scheme 2 Gap |
|---|---|---|---|---|
| *Mix.* (all passages) | 96.7% | 86.7% | — | — |
| *Comp.* (all passages) | 63.3% | 43.3% | 33.3% | 43.3% |
| *Comp.* (excl. 8, 10) | 79.2% | 54.2% | 17.5% | 32.5% |
| *Comp.* (excl. 3, 8, 10) | 85.7% | 61.9% | 11.0% | 24.8% |

*Note: Pass Rate = High Pass + Low Pass. Gap =* Mix. *pass rate minus* Comp. *pass rate under each scheme. Stratification excludes specified* Comp. *passages;* Mix. *includes all passages in all rows.*

The quality rating analysis reveals a dimension of performance not fully captured by mean TQS. While section 4.2 demonstrated that mean TQS converges when catastrophic passages are excluded (gap of 2.1 points when excluding passages 3, 8, and 10), pass rates under the quality rating schemes do not converge to the same degree. Even excluding passages 3, 8, and 10, *Comp*. shows a pass-rate gap of 11.0 percentage points under Scheme 1 (*Mix*. 96.7% vs. *Comp*. 85.7%) and 24.8 percentage points under Scheme 2 (*Mix*. 86.7% vs. *Comp*. 61.9%).

This divergence reflects the distribution of critical errors: the excluded passages account for 86% of all *Comp.* critical errors, although the remaining passages still contain more critical errors per translation than *Mix.* (8 across 21 vs. 5 across 30). Notably, passages 2 and 5 each contain critical errors from all three models despite mean TQS scores of 86.7 and 89.5 respectively (LOW PASS under Scheme 2), indicating that these passages present systematic challenges not captured by the catastrophic-passage framing. The models thus achieve similar *average* quality on expository *Comp*. passages but with *lower reliability*, underscoring that previously untranslated technical texts pose greater risks even outside the most challenging passages, especially for non-expert users unable to verify directly from the AG source.

### 4.5 Error Typology

Table 8 presents the distribution of MQM errors by type and severity across *Mix*. and *Comp*.

**Table 8. MQM Error Typology Distribution by Text**

| | ***Mix*.** | ***Comp*.** | ***Comp*./*Mix*. Ratio** |
|---|---|---|---|
| **Total errors** | 170 | 265 | 1.6× |
| Errors per passage | 5.7 | 8.8 | 1.6× |
| **Severity** | | | |
| Neutral | 29 (17.1%) | 45 (17.0%) | 1.0× |
| Minor | 103 (60.6%) | 105 (39.6%) | 0.7× |
| Major | 33 (19.4%) | 58 (21.9%) | 1.1× |
| Critical | 5 (2.9%) | 57 (21.5%) | 7.4× |
| **Error Type** | | | |
| Terminology | 74 (43.5%) | 198 (74.7%) | 1.7× |
| Accuracy | 96 (56.5%) | 67 (25.3%) | 0.4× |
| **Terminology Subtypes** | | | |
| Term. Accuracy | 60 (35.3%) | 192 (72.5%) | 2.1× |
| Term. Consistency | 14 (8.2%) | 6 (2.3%) | 0.3× |
| **Accuracy Subtypes** | | | |

| Mistranslation | 44 (25.9%) | 39 (14.7%) | 0.6× |
|---|---|---|---|
| Overtranslation | 14 (8.2%) | 6 (2.3%) | 0.3× |
| Undertranslation | 26 (15.3%) | 8 (3.0%) | 0.2× |
| Addition | 8 (4.7%) | 9 (3.4%) | 0.7× |
| Omission | 4 (2.4%) | 5 (1.9%) | 0.8× |

*Note: Percentages are of total errors within each text. Error type percentages may sum to more than 100% because individual errors can be categorized under both Terminology and Accuracy. Total errors across both texts = 435.*

*Comp*. produced 56% more total errors than *Mix*. (265 vs. 170), but the distribution differed qualitatively, not merely quantitatively. The most prominent contrast was in severity: critical errors constituted 21.5% of all *Comp*. errors but only 2.9% of *Mix*. errors, a 7.4× ratio. On *Mix*., errors were overwhelmingly minor (60.6%), reflecting small inaccuracies that did not compromise comprehension.

Error types also diverged sharply. On *Comp*., terminology errors dominated (74.7% of total), nearly all of which were classified as terminological accuracy errors (72.5%) rather than consistency errors (2.3%). On *Mix*., error types were more evenly distributed, with accuracy errors (56.5%)—principally mistranslation (25.9%) and undertranslation (15.3%)—exceeding terminology errors (43.5%). This pattern suggests that on *Mix*., where all three models achieved high TQS, the residual errors tended toward general translational imprecision rather than failures of specialized vocabulary. On *Comp*., by contrast, the difficulty of rendering the technical terminology—pharmacological ingredients, anatomical terms, and extremely rare words—drove both the overall error count and the disproportionate share of critical-severity errors.

## 5. Discussion

### 5.1 Translation Quality: Previously Translated vs. Untranslated Texts

The central question that we posed at the outset of this study—how well do general-purpose LLMs “out of the box” translate ancient technical prose?—requires a nuanced answer based on our findings. On the previously translated text (*Mix*.), LLM performance was strong: mean aggregate TQS of 95.2 with low volatility (SD = 4.8), a 97% pass rate under Scheme 1, and only 5 critical errors across 30 translations. One LLM (Claude) achieved a perfect record on *Mix*., with zero critical errors and 100% of translations rated HIGH PASS or LOW PASS under both quality schemes, and another achieved a near-perfect record (Gemini). These results suggest that for expository Galenic prose with available modern translations, current LLMs can produce serviceable first-draft translations adequate for grasping the key claims and concepts of the source text.

On previously untranslated texts (*Comp*.), performance was substantially weaker: mean aggregate TQS of 79.9 with high volatility (SD = 26.2), a 63% pass rate under Scheme 1 (43% under Scheme 2), and 57 critical errors across 30 translations. However, the aggregate gap of 15.3 TQS points between *Mix*. and *Comp*. obscures important heterogeneity. As demonstrated in section 4.2, performance on Comp. was bimodal rather than uniformly degraded: two densely pharmacological passages (8 and 10) produced catastrophic failures across all models, while the remaining expository passages achieved quality within 2–4 points of *Mix*. It remains to consider what explains the gap between performance on the translated and untranslated texts.

### 5.1.1 The Memorization Hypothesis

One plausible explanation for the *Mix*.–*Comp*. gap is that LLMs have “memorized” human translations. When presented with the AG text, the LLMs may be paraphrasing them via semantic leakage from material observed during training (Ippolito et al. 2023). If this were a determinative factor, performance on *Mix.* could be misleading as to the performance of LLMs in translating AG because of contamination by the pre-training data (1.2).

First, the persistence of the BLEU gap even when the MQM gap disappears provides partial support for the supposition of memorization. On expository *Comp*. passages (excluding 3, 8, 10), MQM metrics converged to within 2.1 points of *Mix*., yet relative drop in performance hardly changed (-47.4% versus -48.6% for all passages, Table 4). If LLMs were translating both texts equivalently, we might expect lexical similarity to reference translations to converge too. This gap suggests that *Mix*. translations share more surface-level phrasing with published references than *Comp*. translations. This result is consistent with, though not proof of, memorization.

Comparison of reference translations further nuances evaluation (Table SM4). When *Mix.* translations were evaluated separately against the two published references, lexical metrics showed marked preference for the Johnston (Loeb) translation (67–73% of comparisons), while COMET preferred Singer and van der Eijk (77%). However, absolute BLEU scores against both references remained in the .13–.36 range, below levels expected from verbatim reproduction (>50%). LLM outputs thus differ substantially on the surface from both translations on the surface. Memorization likely plays some role, but the evidence is not determinative.

Third, however, the distribution of critical errors suggests that the *Mix*.–*Comp*. gap reflects genuine translation difficulty rather than memorization alone. Critical errors on *Comp*. were concentrated in passages with dense technical terminology (esp. 8 and 10). The error typology analysis further revealed that 75% of *Comp*. errors were terminological compared to 44% on *Mix* (section 4.5). Scattered critical failures negatively impacting *Comp.* pass rates under both quality evaluation schemes across all passages also reflect persistent difficulty with terminology. If memorization were the sole explanation for superior *Mix*. performance, we would not expect such a clear content-based pattern in *Comp*. failures.

### 5.1.2 Difficult Technical Terminology Is a Predictor of Failure

The evidence thus points towards the difficulty of technical terminology, and not only prior translation availability, as a significant predictor of LLM translation failure. Empirical review of the catastrophically failed passages 8 and 10 as well as the remaining 17 critical errors in other *Comp.* passages helps to determine the salient "difficulty" that led LLMs to mistranslate the source. Qualitatively, failures clustered around pharmacological terminology including mineral compounds (20/57), plant names (15/57), and specialized descriptions for drug properties (7/57). Some of the terms on which LLMs failed present genuine stumbling blocks for human translators, insofar as the meaning of the Greek term is intrinsically uncertain (e.g. χαλκῖτις**,** χάλκανθος**,** λιθαργύρου χρυσῖτις**,** ἀμπελῖτις) But many highly technical terms on which LLMs failed pose little difficulty for expert human translators, in that they have unambiguous English equivalents (e.g. μῆον, οἴσυπος, χάρτης, ἐλελίσφακος, μελάνθιον, μάλαγμα); the LLMs nonetheless produced incorrect or only partially correct translations or uninformative or misleading transliterations. Since LLMs do not fail in translating technical terms of all kinds (such as the more common *dunamis*, "capacity," or *krasis*, "mixture"), the pattern suggests specifically that it is only particularly recondite or abstruse terminology on which the machines consistently struggle.

To substantiate this conclusion, we computed corpus-based metrics for each Galenic passage using the Diorisis Ancient Greek Corpus (Vatri and McGillivray 2018), a 10-million-word-token corpus containing morphologically enriched data including lemmatized forms for canonical AG texts from Homer to the fifth century CE. We lemmatized chunks of *Mix.* and *Comp.* using Stanza (Qi et al. 2020) and calculated the proportions of lemmas that were either absent from Diorisis or appeared fewer than 50 times, which we aggregated to establish a "rare terms ratio," as well as the mean Zipf-scaled frequency across all lemmata. Since Diorisis is a principally literary not technical corpus, *difficulty* is thus handled as a function of *rarity* in non-technical language. These data are presented in full in the Supplementary Material (Table SM5). Some summary observations are presented here with respect to *Comp.*:

- Mean rare term ratio was 23.5% for *Comp.* against 14% for *Mix.*
- The two catastrophically failed passages had rare term ratios of .407 (8) and 0.434 (10), while no other passage had a rare term ratio of >0.24 (none in *Mix.* had rare term ratio >0.188). The mean rare term ratio of these two passages (0.421) is more than double that of the other *Comp.* passages (0.189), with mean TQS of 33.8 versus 91.4.
- For passages in *Comp.* with rare term ratio <0.193, no models failed under Scheme 1 and only one model failed under Scheme 2 (Claude, passage 4, but TQS at a respectable 89.8, LOW PASS).

Table 9 presents correlations between the difficulty metrics and MQM scores.

**Table 9. Correlations between Terminology Rarity Metrics and Translation Quality**

| **Predictor** | **TQS (*Comp.*)** | **TQS (All)** | **Critical (*Comp.*)** | **Critical (All)** |
|---|---|---|---|---|
| Rare term ratio | −.97*** | −.93*** | +.92*** | +.90*** |
| Not-found ratio | −.96*** | −.94*** | +.91*** | +.91*** |
| Average Zipf frequency | +.96*** | +.95*** | −.91*** | −.92*** |

*Note: Pearson* r *values.* Comp. = Comp. *passages only (*n *= 10); all = all passages (*n *= 20). Rare term ratio = proportion of lemmas with Diorisis frequency < 50 or not found. Not-found ratio = proportion of lemmas absent from Diorisis. Average Zipf frequency = mean Zipf-scaled corpus frequency across passage lemmas (higher = more common vocabulary). **** p *< .001.*

For *Comp.*, the rare term ratio was strongly correlated with TQS ($r = -.97$, $p < .001$, $n = 10$) as well as with critical error count ($r = +.92$, $p < .001$). A simple linear regression showed that rare term ratio explained 93.9% of variance in passage-level TQS ($R^2 = .94$; bootstrap 95% CI: .42–.99). However, Cook's distance analysis identified passages 8 and 10 as highly influential observations ($D$ = 1.50 and 2.03, respectively, against a 4/n threshold of 0.40), and excluding these two terminologically extreme passages substantially attenuated the relationship ($R^2 = .50$, $r = -.71$, $p = .051$). The strong full-range correlation thus reflects the complete spectrum of terminological difficulty, including extreme cases, and should not be interpreted as implying that small increments in rarity produce proportional quality declines. Be that as it may, a Spearman rank correlation—which is less sensitive to extreme values than Pearson *r*—confirmed that the monotonic relationship remained robust across all ten passages ($\rho = -.88$, $p < .001$). Taken together, the quantitative analysis even on this small sample thus underscores that rarity with respect to AG literary writing tracks with degradation of LLM translation performance, a finding that adds further weight to the underlying philological analysis of the critical errors in *Comp.* highlighted above.

The strong predictive power of corpus frequency likely reflects a general property of LLM competence: model performance degrades as tasks move away from the statistical center of the training distribution. Because the digitized corpus of AG likely ingested during pre-training—dominated by open-access repositories like the Perseus Digital Library—is largely literary and canonical in nature, the Diorisis corpus acts as an effective proxy for "in-distribution" data of the LLMs. Terms that are rare in Diorisis are therefore statistically marginal in the training data, forcing models to operate at the boundaries of learned representations. The high density of rare terms in the failed *Comp.* passages thus forces the models to operate at the margins of their training data. Lacking a robust statistical foundation for these specific technical tokens, the models are prone to fill the gap with the catastrophic hallucinations observed in our study. This mechanism is not specific to AG: the same principle predicts that LLMs will fail on rare terminology in any specialized domain where the training data overrepresents general-purpose text relative to domain-specific material.

### 5.2 Evaluating the Evaluators

A secondary aim of this study was to benchmark standard automated MT evaluation metrics against expert human judgment to determine the utility of these metrics. The results reveal significant limitations of automated metrics for evaluating LLM translation of AG, with important implications for future research in this domain.

#### 5.2.1 Correlations are Text Dependent

Table 5 demonstrates that several evaluation metrics showed moderate or (in the case of BERTScore) strong correlations with TQS. When correlations are computed separately by text, a striking finding emerges nuancing these results. Table 10 presents this breakdown.

**Table 10. Correlations between Automated Metrics and MQM TQS by Text**

| Metric | ***Mix*. (n=30)** | ***Comp*. (n=30)** | **Combined (n=60)** |
|---|---|---|---|
| BERTScore | −0.10 | +0.85*** | +0.75*** |
| COMET | −0.07 | +0.62*** | +0.60*** |
| METEOR | +0.02 | +0.55** | +0.53*** |
| chrF++ | −0.00 | +0.55** | +0.53*** |
| BLEU-4 | −0.08 | +0.42* | +0.45*** |
| ROUGE-L | +0.13 | +0.24 | +0.34** |
| BLEURT | +0.00 | +0.18 | +0.32* |

*Note: Pearson* r *values. ** p *< .05, ** p < .01, *** p < .001.* Mix. *translations cluster in narrow high-quality band (TQS SD = 4.8);* Comp. *translations show high variance (TQS SD = 26.2).*

On *Mix*., where translations clustered in a narrow high-quality band (TQS M = 95.2, SD = 4.8), no automated evaluations correlated significantly with TQS. On *Comp*., where translation quality varied widely (TQS M = 79.9, SD = 26.2), more metrics showed moderate-to-strong correlations, with BERTScore achieving $r = .85$. Again, the ordering of metrics here is merely suggestive, since confidence intervals for adjacent metrics overlap at this sample size, but the important result is that the aggregate correlations reported in Table 5 are driven entirely by *Comp*. An important conclusion is thus that automated metrics track variance in translation quality but cannot discriminate among good translations.

This finding has methodological implications. Studies evaluating MT on well-translated texts may find that automated metrics show little relationship to human judgment because there is insufficient quality variance to predict. Conversely, metrics may appear more reliable than they are when evaluated on corpora containing a mix of good and catastrophically bad translations, as the catastrophic cases drive the correlation.

### 5.2.2 Automatic Evaluation Versus Human Judgement

What can be said about the efficacy of the automated metrics in evaluating translation quality? With limitations owing to our small sample size in mind, some general observations can be ventured: BERTScore showed the strongest correlation with TQS ($r$ = .75, 95% CI: .61–.84), clearly outperforming ROUGE-L and BLEURT ($r$ = .34 and .32). The remaining metrics—COMET, METEOR, chrF++, and BLEU-4—fell in an intermediate range ($r$ = .45–.60) that could not be reliably differentiated from one another at this sample size, although see the qualifications below for COMET.

The greater morphological sensitivity of chrF++ and METEOR yielded at best marginal advantages over BLEU-4. BERTScore's superior performance relative to BLEURT may suggest that for domain-shifted evaluation, simpler embedding-based similarity outperforms learned metrics, although COMET's relatively strong showing means this remains tentative.

While BERTScore appears to be strongest single evaluation metric, our results make it clear that no automated metrics should be trusted in isolation, particularly since the strength of correlation depends closely on the spread of translation quality, a parameter unlikely to be known *a priori*. Supplementing the Pearson correlations by text in Table 10 with Spearman rank correlations nuances the picture further, revealing that only BERTScore and Comet achieved significance ($\rho$ = +0.71 and +0.65 respectively at $p < .001$). Even within *Comp.*, then, only these two metrics demonstrated a genuinely monotonic relationship with TQS throughout their distribution, other metrics being skewed by the extreme outlier passages (8 and 10). Thus, even if automated metrics are used as efficient screening tools—in particular, flagging translations for closer review when scores fall below thresholds—they cannot substitute for expert philological analysis.

Some further limitations emerged. While all automated metrics compressed the quality range relative to human judgment, it is notable that even the neural metric COMET masked catastrophic failures. Human evaluation revealed a 56.7-point quality drop (TQS) between the catastrophic (8, 10) and other passages, but COMET compressed this into a 4.8% dip (77.3% vs. 72.5%). A practitioner relying on COMET scores would be seriously misled on the most terminologically difficult passages, including:

- *Comp*. 8 ChatGPT: TQS = 0.0, COMET = 73.3%
- *Comp*. 10 Gemini: TQS = 25.2, COMET = 75.6%
- *Comp*. 8 Claude: TQS = 21.8, COMET = 74.0%

Conversely, BLEU systematically underrated high-quality translations that diverged lexically from reference texts:

- *Comp*. 6 ChatGPT: TQS = 95.7, BLEU = 9.1%
- *Comp*. 6 Claude: TQS = 97.9, BLEU = 13.8%
- *Comp*. 7 ChatGPT: TQS = 97.8, BLEU = 17.6%

These translations achieved near-perfect human evaluation scores but received failing BLEU grades (<.20). The discrepancy reflects BLEU's reliance on exact n-gram matching, which suits poorly with AG's interpretative plurality, even if the problematic appearance of the metric may be exacerbated by the lack of multiple reference translations for *Comp.*

## 6 Conclusion

This study provides the first systematic, reference-free human evaluation of LLM-based machine translation for Ancient Greek technical prose, benchmarking three commercial LLMs against expert philological judgment using a modified MQM framework. Our findings suggest that current LLMs produce serviceable translations of expository Galenic prose—whether previously translated or not—but struggle with technical content dense with terminology rare in canonical AG literature. In such passages, critical errors are common even in otherwise adequate translations. Beyond Classics, our findings thus suggest a practical heuristic for any domain considering LLM deployment on specialized material: when the task involves terminology that is rare or absent from general-purpose corpora, model output should be treated as provisional regardless of apparent fluency. Corpus frequency analysis offers a cheap, scalable method for flagging content likely to contain critical errors prior to expert review, potentially enabling efficient quality assurance for large-scale LLM-assisted analysis pipelines.

We found further that automated MT evaluation metrics show moderate correlation with human judgment but cannot be trusted in isolation: BERTScore outperforms ROUGE-L and BLEURT, but other metrics cannot be teased apart at this sample size. Additionally, metrics only demonstrated significant correlations on *Comp.*, a text with a wide translation quality spread. In short, automated evaluation metrics cannot substitute for philological judgment.

Based on our findings, we further offer the following conclusions tailored to different readerships of Galenic texts:

*For expert users (Classics or historians of philosophy, medicine, or science with working knowledge of AG):*

- Current LLMs produce serviceable first-draft translations of expository Galenic prose regardless of whether prior human translations exist: note mean TQS ≥ 91 and 86% pass rate (under the more permissive Scheme 1) for non-catastrophic passages, with quality approaching expert level for common vocabulary (in our dataset, rare term ratio <0.20).
- LLMs are reliably useful for syntactic parsing and initial orientation to unfamiliar texts; general sense and argument structure are typically preserved even when individual terms are mistranslated (as with the scattered critical errors in *Comp.*).
- Critical errors concentrate around specialized technical terminology including anatomical parts, pathological conditions, and properties of substances that are rarely attested in or absent from the AG literary corpus. In our dataset, passages with rare term ratios above .30 produced catastrophic failures across all three models. Expert users should treat technical vocabulary in LLM output as provisional and verify against the source.
- LLMs struggle with textual corruption and may confidently render nonsense; the fluency of neural MT can mask underlying incoherence that a human translator would flag.

*For non-expert users (students, scholars outside Classics, general readers):*

- LLM translations of expository Ancient Greek prose are broadly reliable for grasping the main claims and concepts of a text, but any passage with specialized technical terminology should be treated with caution: fluent output can mask serious errors invisible without source-language competence.
- Under Scheme 2, our conservative quality rating scheme (which flags any translation with even one critical error), only 43% of translations from a previously untranslated pharmacological text passed, compared to 87% for a previously translated expository text.
- When using LLM translations for research or study, non-experts should cross-check claims involving specialized vocabulary (substances, anatomical terms, pathological conditions) against secondary scholarship and should not rely on a single LLM translation for evidential purposes.

Our findings open several paths for further research. First, the qualitative observations compiled during MQM scoring—not fully captured in the quantitative error typology—could support granular characterization of LLM failure modes, particularly fluent hallucination in the presence of textual corruption or challenging terminology. Such analysis could yield pragmatic heuristics for identifying and correcting MT errors.

Second, the error patterns observed here suggest opportunities for prompt optimization. LLMs could be instructed to transliterate rather than translate technical terms of uncertain meaning (adding a "best guess" in brackets for example)—a strategy already employed by expert human translators of AG technical prose. Systematic exploration of such interventions could substantially improve MT reliability for specialized texts.

Third, the automated metrics benchmarked here could now be applied—with appropriate caution—to extend these findings across Galen's corpus and beyond. MQM evaluation at scale is impractical, but targeted spot-checking guided by metric thresholds could enable efficient quality assurance for larger MT projects. Extension of the corpus would also provide sufficient statistical power to test the observed relationship between terminology rarity and translation quality with greater precision than the present sample permits.

Fourth, it remains uncertain how MQM scores map onto the broader distribution of expert preferences, given interpretive plurality in AG translation. A complementary evaluation based on blind expert comparison between human and MT translations could illuminate dimensions of quality not captured by error-based assessment.

## 7. Limitations

We note the following limitations:

1. **Iteration consistency**. We did not explore consistency of model performance across repeated iterations of the MT task.
2. **Cross-passage consistency**. We evaluated intra-passage accuracy but not cross-passage terminological consistency. Utility of LLM-based MT could degrade without controlling for consistency across multiple inputs.

3. **Generalizability**. The dataset was selected to represent a spectrum of content types but does not fully capture the diversity of Galen's corpus or AG technical prose more broadly.
4. **Model evolution**. We used state-of-the-art model versions available at the time of data collection, but LLM capabilities are advancing rapidly and our findings represent a snapshot that may not reflect current or future performance.
5. **Single evaluator consensus**. While MQM scoring involved multiple expert reviewers reaching 100% consensus, the final judgments reflect one team's interpretive standards.
6. **Reference translation availability for *Comp.*** Automated metrics for Comp. relied on a single expert-created reference translation, whereas *Mix*. had two published references. This asymmetry may have inflated the apparent *Mix*.–*Comp*. gap on reference-dependent metrics.
7. **Statistical power for passage-level analyses**. Passage-level correlations between terminology difficulty metrics and translation quality are based on 10 passages per text. While the observed correlations are strong and significant, quantitative findings should be understood as characterizing the pattern observed in this dataset rather than as precise parameter estimates.

## 8. Appendix

### A1. Prompt for LLM Translation

"""You are a Classical philologist specializing in Ancient Greek. Translate the following Ancient Greek text to English.

Context: This is ancient Greek text. Please:
- Preserve technical terminology and scholarly precision
- Maintain the academic tone of the original
- Provide clear, readable English while respecting the ancient context
- Pay attention to classical Greek grammar and syntax

Provide only the English translation, no explanations.

Ancient Greek text:
{text}"""

---

[i] All data and code are available at https://github.com/campattison/galen_project.

## Supplementary Material: Tables 1–5

### Table SM1. AG dataset

| Work | # | Location | Words |
|---|---|---|---|
| *Mix.* | 1 | I.1, 1,1–2,3 H.; I.509,1–510,15 K. | 215 |
| | 2 | I.4, 12,13–13,10 H.; I.527,13–529,2 K. | 211 |
| | 3 | I.6, 23,16–24,10 H.; I.545,14–547,4 K. | 203 |
| | 4 | I.9, 35,3–28 H.; I.564,7–565,14 K. | 222 |
| | 5 | II.2, 47,22–48,16 H.; I.584,3–585,10 K. | 210 |
| | 6 | II.3, 59,6–60,4 H.; I.602,11–604,1 K. | 223 |
| | 7 | II.5–6, 70,6–71,5 H.; I.620,13–622,5 K. | 231 |
| | 8 | II.6, 81,10–82,3 H.; I.638,11–639,16 K. | 206 |
| | 9 | III.2, 93,21–94,21 H.; I.658,5–659,13 K. | 232 |
| | 10 | III.4, 105,2–27 H.; I.676,17–678,9 K. | 229 |
| *Comp.* | 1 | I.2, XIII.367,10–369,1 K. | 224 |
| | 2 | I.15–16, XIII.431,4–432,13 K. | 224 |
| | 3 | II.4–5, XIII.499,5–500,15 K. | 233 |
| | 4 | III.2, XIII.569,10–570,17 K. | 228 |
| | 5 | III.7–8, XIII.635,6–636,14 K. | 216 |
| | 6 | IV.5, XIII.703,8–704,13 K. | 200 |
| | 7 | V.2, XIII.771,18–773,9 K. | 232 |
| | 8 | V.13, XIII.840,10–842,1 K. | 222 |
| | 9 | VI.10, XIII.909,17–911,7 K. | 223 |
| | 10 | VII.7, XIII.983,3–984,8 K. | 225 |
| | | **Average Word Count** | **220.5** |

*Note: References to Galen's* On Mixtures *follow the page and line numbers in the critical edition used as the basis for the reference translations – G. Helmreich (H.),* Claudius Galenus. De temperamentis libri III *(Helmreich 1904) – as well as the corresponding volume, page, and line numbers in the Kühn edition (K.) (Kühn 1827). References to Galen's* On Composition of Drugs According to Kinds *follow Kühn, the only available edition. The traditional division of the text into chapters has also been preserved.*

### Table SM2. Selected Automated Metric Scores for All 60 LLM Translations

| Text | Chunk | Model | TQS | BLEU-4 | BERTScore | COMET | Rating (Scheme 1) |
|---|---|---|---|---|---|---|---|
| ***Mix.*** | | | | | | | |
| | 1 | Claude | 96.7 | 45.7% | 92.9% | 82.6% | HP |
| | 1 | Gemini | 99.3 | 41.1% | 93.4% | 83.1% | HP |
| | 1 | ChatGPT | 97.0 | 39.5% | 92.7% | 82.4% | HP |
| | 2 | Claude | 92.0 | 36.1% | 91.9% | 83.0% | LP |
| | 2 | Gemini | 99.7 | 32.5% | 91.0% | 83.2% | HP |
| | 2 | ChatGPT | 77.8 | 34.4% | 91.0% | 82.8% | F |
| | 3 | Claude | 100.0 | 24.4% | 90.8% | 79.2% | HP |
| | 3 | Gemini | 100.0 | 30.1% | 90.3% | 81.2% | HP |
| | 3 | ChatGPT | 97.5 | 24.2% | 90.3% | 80.7% | HP |
| | 4 | Claude | 96.5 | 31.9% | 90.2% | 77.2% | HP |
| | 4 | Gemini | 99.1 | 32.0% | 90.9% | 78.8% | HP |
| | 4 | ChatGPT | 96.7 | 29.5% | 89.6% | 78.8% | HP |
| | 5 | Claude | 97.0 | 32.2% | 91.1% | 77.4% | HP |
| | 5 | Gemini | 97.1 | 32.2% | 91.1% | 77.6% | HP |
| | 5 | ChatGPT | 89.6 | 24.6% | 90.1% | 76.5% | LP |
| | 6 | Claude | 98.0 | 42.8% | 93.3% | 79.7% | HP |
| | 6 | Gemini | 90.1 | 42.8% | 93.2% | 80.8% | LP |
| | 6 | ChatGPT | 87.6 | 42.2% | 93.1% | 80.9% | LP |
| | 7 | Claude | 97.1 | 33.6% | 92.1% | 81.6% | HP |
| | 7 | Gemini | 97.5 | 38.8% | 91.7% | 81.4% | HP |
| | 7 | ChatGPT | 97.9 | 32.3% | 91.0% | 80.5% | HP |
| | 8 | Claude | 94.8 | 33.2% | 91.4% | 78.0% | LP |
| | 8 | Gemini | 98.9 | 34.5% | 91.2% | 79.5% | HP |
| | 8 | ChatGPT | 97.9 | 27.6% | 89.8% | 78.4% | HP |
| | 9 | Claude | 93.3 | 33.1% | 91.2% | 80.1% | LP |
| | 9 | Gemini | 92.7 | 29.5% | 91.6% | 80.2% | LP |
| | 9 | ChatGPT | 93.1 | 33.6% | 91.5% | 79.7% | LP |
| | 10 | Claude | 97.1 | 29.0% | 91.1% | 78.8% | HP |
| | 10 | Gemini | 96.7 | 28.5% | 91.0% | 80.9% | HP |
| | 10 | ChatGPT | 88.1 | 26.4% | 91.1% | 78.3% | LP |
| ***Comp.*** | | | | | | | |
| | 1 | Claude | 100.0 | 24.8% | 91.2% | 79.3% | HP |
| | 1 | Gemini | 99.4 | 20.0% | 91.0% | 80.1% | HP |
| | 1 | ChatGPT | 99.3 | 19.3% | 90.5% | 78.0% | HP |
| | 2 | Claude | 88.8 | 19.4% | 90.4% | 78.5% | LP |
| | 2 | Gemini | 85.4 | 26.4% | 91.3% | 79.8% | F |
| | 2 | ChatGPT | 86.0 | 21.6% | 89.6% | 75.8% | F |
| | 3 | Claude | 83.0 | 16.2% | 89.8% | 75.8% | F |
| | 3 | Gemini | 88.3 | 14.9% | 89.2% | 72.9% | LP |
| | 3 | ChatGPT | 65.9 | 10.9% | 89.2% | 75.2% | F |
| | 4 | Claude | 89.8 | 17.2% | 91.0% | 75.6% | LP |
| | 4 | Gemini | 95.2 | 23.2% | 91.0% | 76.6% | HP |
| | 4 | ChatGPT | 97.1 | 21.9% | 90.8% | 75.3% | HP |
| | 5 | Claude | 89.7 | 16.8% | 90.4% | 77.0% | LP |
| | 5 | Gemini | 90.5 | 14.9% | 90.0% | 77.7% | LP |
| | 5 | ChatGPT | 88.4 | 16.9% | 90.0% | 77.2% | LP |
| | 6 | Claude | 97.9 | 13.8% | 90.5% | 76.7% | HP |
| | 6 | Gemini | 95.8 | 21.4% | 91.1% | 78.7% | HP |
| | 6 | ChatGPT | 95.7 | 9.1% | 90.2% | 75.7% | HP |
| | 7 | Claude | 94.2 | 18.8% | 90.0% | 79.9% | LP |
| | 7 | Gemini | 93.9 | 13.9% | 89.0% | 78.7% | LP |
| | 7 | ChatGPT | 97.8 | 17.6% | 89.7% | 78.0% | HP |
| | 8 | Claude | 21.8 | 19.4% | 86.8% | 74.0% | F |
| | 8 | Gemini | 65.8 | 18.0% | 87.2% | 74.7% | F |
| | 8 | ChatGPT | 0.0 | 6.4% | 85.3% | 73.3% | F |
| | 9 | Claude | 92.8 | 13.6% | 89.6% | 75.9% | LP |
| | 9 | Gemini | 95.0 | 19.7% | 90.3% | 77.9% | LP |
| | 9 | ChatGPT | 82.8 | 18.9% | 90.5% | 77.6% | F |
| | 10 | Claude | 54.1 | 6.6% | 86.7% | 72.5% | F |
| | 10 | Gemini | 25.3 | 17.6% | 88.7% | 75.6% | F |
| | 10 | ChatGPT | 36.0 | 14.2% | 85.0% | 64.6% | F |

*Note: BLEU-4, BERTScore, and COMET scores shown as percentages. For* Mix.*, multi-reference scores are reported (maximum across Johnston and Singer and van der Eijk references). For* Comp.*, single expert reference translation used. TQS and Rating from MQM human evaluation included for comparison. Rating = Scheme 1 (HP = High Pass, LP = Low Pass, F = Fail).*

## Table SM3. Per-Passage MQM Scores for All 60 LLM Translations

| Text | Ch. | Model | TQS | Rating | Crit.? | N | Mi | Ma | Cr |
|---|---|---|---|---|---|---|---|---|---|
| ***Mix.*** | | | | | | | | | |
| | 1 | Claude | 96.7 | HP | No | 0 | 4 | 1 | 0 |
| | 1 | Gemini | 99.3 | HP | No | 0 | 2 | 0 | 0 |
| | 1 | ChatGPT | 97.0 | HP | No | 0 | 3 | 1 | 0 |
| | 2 | Claude | 92.0 | LP | No | 1 | 1 | 4 | 0 |
| | 2 | Gemini | 99.7 | HP | No | 1 | 1 | 0 | 0 |
| | 2 | ChatGPT | 77.8 | F | Yes | 1 | 3 | 1 | 2 |
| | 3 | Claude | 100.0 | HP | No | 0 | 0 | 0 | 0 |
| | 3 | Gemini | 100.0 | HP | No | 0 | 0 | 0 | 0 |
| | 3 | ChatGPT | 97.5 | HP | No | 0 | 2 | 1 | 0 |
| | 4 | Claude | 96.1 | HP | No | 1 | 6 | 1 | 0 |
| | 4 | Gemini | 99.1 | HP | No | 0 | 3 | 0 | 0 |
| | 4 | ChatGPT | 96.7 | HP | No | 0 | 4 | 1 | 0 |
| | 5 | Claude | 97.0 | HP | No | 1 | 8 | 0 | 0 |
| | 5 | Gemini | 97.1 | HP | No | 4 | 3 | 1 | 0 |
| | 5 | ChatGPT | 89.6 | LP | No | 2 | 7 | 4 | 0 |
| | 6 | Claude | 98.0 | HP | No | 0 | 5 | 0 | 0 |
| | 6 | Gemini | 90.1 | LP | Yes | 1 | 2 | 0 | 1 |
| | 6 | ChatGPT | 87.6 | LP | Yes | 1 | 2 | 1 | 1 |
| | 7 | Claude | 97.1 | HP | No | 1 | 3 | 1 | 0 |
| | 7 | Gemini | 97.5 | HP | No | 0 | 3 | 1 | 0 |
| | 7 | ChatGPT | 97.5 | HP | No | 1 | 7 | 0 | 0 |
| | 8 | Claude | 94.8 | LP | No | 1 | 4 | 2 | 0 |
| | 8 | Gemini | 98.9 | HP | No | 1 | 3 | 0 | 0 |
| | 8 | ChatGPT | 97.9 | HP | No | 2 | 0 | 1 | 0 |
| | 9 | Claude | 93.3 | LP | No | 1 | 4 | 3 | 0 |
| | 9 | Gemini | 92.7 | LP | No | 0 | 8 | 3 | 0 |
| | 9 | ChatGPT | 93.1 | LP | No | 2 | 5 | 3 | 0 |
| | 10 | Claude | 97.1 | HP | No | 3 | 3 | 1 | 0 |
| | 10 | Gemini | 96.7 | HP | No | 4 | 5 | 1 | 0 |
| | 10 | ChatGPT | 88.1 | LP | Yes | 0 | 2 | 1 | 1 |
| ***Comp.*** | | | | | | | | | |
| | 1 | Claude | 100.0 | HP | No | 4 | 0 | 0 | 0 |
| | 1 | Gemini | 99.4 | HP | No | 5 | 2 | 0 | 0 |
| | 1 | ChatGPT | 99.3 | HP | No | 6 | 2 | 0 | 0 |
| | 2 | Claude | 88.8 | LP | Yes | 2 | 4 | 1 | 1 |
| | 2 | Gemini | 85.4 | F | Yes | 1 | 1 | 4 | 1 |
| | 2 | ChatGPT | 86.0 | F | Yes | 1 | 5 | 2 | 1 |
| | 3 | Claude | 83.0 | F | Yes | 0 | 3 | 5 | 1 |
| | 3 | Gemini | 88.3 | LP | Yes | 0 | 4 | 2 | 1 |
| | 3 | ChatGPT | 65.9 | F | Yes | 0 | 1 | 1 | 4 |
| | 4 | Claude | 89.8 | LP | Yes | 0 | 4 | 0 | 1 |
| | 4 | Gemini | 95.2 | HP | No | 2 | 5 | 2 | 0 |
| | 4 | ChatGPT | 97.1 | HP | No | 0 | 3 | 1 | 0 |
| | 5 | Claude | 89.7 | LP | Yes | 1 | 2 | 1 | 1 |
| | 5 | Gemini | 90.5 | LP | Yes | 2 | 3 | 1 | 1 |
| | 5 | ChatGPT | 88.4 | LP | Yes | 3 | 2 | 2 | 1 |
| | 6 | Claude | 97.9 | HP | No | 0 | 0 | 1 | 0 |
| | 6 | Gemini | 95.8 | HP | No | 1 | 1 | 2 | 0 |
| | 6 | ChatGPT | 95.7 | HP | No | 0 | 5 | 1 | 0 |
| | 7 | Claude | 94.2 | LP | No | 0 | 1 | 3 | 0 |
| | 7 | Gemini | 93.9 | LP | No | 0 | 4 | 3 | 0 |
| | 7 | ChatGPT | 97.8 | HP | No | 1 | 1 | 1 | 0 |
| | 8 | Claude | 21.8 | F | Yes | 3 | 8 | 0 | 9 |
| | 8 | Gemini | 65.8 | F | Yes | 3 | 5 | 2 | 4 |
| | 8 | ChatGPT | 0.0 | F | Yes | 3 | 15 | 2 | 13 |
| | 9 | Claude | 92.8 | LP | No | 0 | 4 | 3 | 0 |
| | 9 | Gemini | 94.9 | LP | No | 0 | 4 | 2 | 0 |
| | 9 | ChatGPT | 82.8 | F | Yes | 1 | 6 | 3 | 1 |
| | 10 | Claude | 54.2 | F | Yes | 2 | 3 | 2 | 5 |
| | 10 | Gemini | 25.2 | F | Yes | 2 | 5 | 4 | 8 |
| | 10 | ChatGPT | 36.0 | F | Yes | 2 | 2 | 7 | 4 |

*Note: TQS = Translation Quality Score (0–100). Rating under Scheme 1: HP = High Pass (≥95), LP = Low Pass (87–94), F = Fail (<87). Crit.? = contains ≥1 critical error. Severity columns: N = Neutral, Mi = Minor, Ma = Major, Cr = Critical.*

**Table SM4. Reference Translation Preferences Across All Metrics**

| Metric | Johnston Preference | Johnston % | Singer & v.d.E. Preference | Singer & v.d.E. % | Total |
|---|---|---|---|---|---|
| BLEU | 21 | 70.0% | 9 | 30.0% | 30 |
| chrF++ | 22 | 73.3% | 8 | 26.7% | 30 |
| METEOR | 20 | 66.7% | 10 | 33.3% | 30 |
| ROUGE-L | 21 | 70.0% | 9 | 30.0% | 30 |
| BERTScore | 17 | 56.7% | 13 | 43.3% | 30 |
| BLEURT | 18 | 60.0% | 12 | 40.0% | 30 |
| COMET | 7 | 23.3% | 23 | 76.7% | 30 |

*Note: Percentages indicate proportion of 30 passages (10 passages × 3 models) where each reference translation scored higher on the given metric. Six of seven metrics prefer the Johnston reference, with only COMET showing strong preference for Singer & van der Eijk.*

**Table SM5. Terminology Rarity Metrics by Passage**

| Text | Ch. | Terms | Avg. Zipf | Rare Ratio | Rare | Not Found | NF Ratio |
|---|---|---|---|---|---|---|---|
| ***Mix.*** | | | | | | | |
| | 1 | 216 | 5.55 | 15.3% | 33 | 27 | 12.5% |
| | 2 | 211 | 5.58 | 13.7% | 29 | 25 | 11.8% |
| | 3 | 204 | 5.66 | 11.8% | 24 | 21 | 10.3% |
| | 4 | 222 | 5.85 | 10.4% | 23 | 17 | 7.7% |
| | 5 | 210 | 5.62 | 15.7% | 33 | 25 | 11.9% |
| | 6 | 224 | 5.54 | 15.2% | 34 | 30 | 13.4% |
| | 7 | 232 | 5.57 | 15.5% | 36 | 26 | 11.2% |
| | 8 | 207 | 5.60 | 14.0% | 29 | 24 | 11.6% |
| | 9 | 233 | 5.81 | 9.4% | 22 | 19 | 8.2% |
| | 10 | 229 | 5.36 | 18.8% | 43 | 34 | 14.8% |
| ***Comp.*** | | | | | | | |
| | 1 | 214 | 5.60 | 14.9% | 32 | 26 | 12.2% |
| | 2 | 228 | 5.16 | 23.3% | 53 | 36 | 15.8% |
| | 3 | 235 | 5.37 | 20.0% | 47 | 31 | 13.2% |
| | 4 | 228 | 5.61 | 17.1% | 39 | 28 | 12.3% |
| | 5 | 218 | 5.38 | 19.3% | 42 | 31 | 14.2% |
| | 6 | 200 | 5.66 | 17.0% | 34 | 23 | 11.5% |
| | 7 | 233 | 5.61 | 19.3% | 45 | 24 | 10.3% |
| | 8 | 232 | 3.84 | 39.7% | 92 | 69 | 29.7% |
| | 9 | 224 | 5.10 | 20.5% | 46 | 36 | 16.1% |
| | 10 | 228 | 3.45 | 43.4% | 99 | 76 | 33.3% |
| | | | | | | | |
| ***Mix.*** | **Mean** | — | 5.61 | 14.0% | — | — | 11.3% |
| ***Comp.*** | **Mean** | — | 5.08 | 23.5% | — | — | 16.9% |

*Note: Terminology metrics computed using the Diorisis Ancient Greek Corpus (Vatri and McGillivray 2018). Terms = total lemmas in passage. Avg. Zipf = mean Zipf-scaled corpus frequency across all lemmas, with terms not found in Diorisis assigned Zipf = 0 (higher values indicate more common vocabulary). Rare ratio = proportion of lemmas with Diorisis frequency < 50 or absent from corpus. Rare = count of rare lemmas. Not Found = count of lemmas absent from Diorisis. NF Ratio = proportion of lemmas not found.*